%% file: main.tex
\documentclass{article} 
\usepackage[numbers]{natbib}
\usepackage[final]{neurips_2025}

\usepackage{microtype}
\usepackage{hyperref}
\usepackage{url}
\usepackage{booktabs}
\usepackage{multirow}
\usepackage{amsthm}
\usepackage{algorithm}
\usepackage{algpseudocode}
\usepackage{amssymb} 
\usepackage{amsmath}
\usepackage{listings}
\usepackage[table]{xcolor}
\usepackage{tikz}
\usetikzlibrary{arrows,arrows.meta,decorations.pathmorphing,backgrounds,positioning,fit,petri,calc,math}

\input{alisa_macros}

\title{Broken Tokens? Your Language Model can\\Secretly Handle \tikz[baseline=-0.6ex] \node[rectangle,rounded corners=4pt,text height=1.5ex,text depth=.25ex,minimum width=0.5cm,minimum height=0.5cm,inner sep=2pt,very thick,draw,font=\ttfamily] () {No};
\tikz[baseline=-0.6ex] \node[rectangle,rounded corners=4pt,text height=1.5ex,text depth=.25ex,minimum width=0.5cm,minimum height=0.5cm,inner sep=2pt,very thick,draw,font=\ttfamily] () {n};
\tikz[baseline=-0.6ex] \node[rectangle,rounded corners=4pt,text height=1.5ex,text depth=.25ex,minimum width=0.5cm,minimum height=0.5cm,inner sep=2pt,very thick,draw,font=\ttfamily] () {-};
\tikz[baseline=-0.6ex] \node[rectangle,rounded corners=4pt,text height=1.5ex,text depth=.25ex,minimum width=0.5cm,minimum height=0.5cm,inner sep=2pt,very thick,draw,font=\ttfamily] () {Ca}; 
\tikz[baseline=-0.6ex] \node[rectangle,rounded corners=4pt,text height=1.5ex,text depth=.25ex,minimum width=0.5cm,minimum height=0.5cm,inner sep=2pt,very thick,draw,font=\ttfamily] () {noni};
\tikz[baseline=-0.6ex] \node[rectangle,rounded corners=4pt,text height=1.5ex,text depth=.25ex,minimum width=0.5cm,minimum height=0.5cm,inner sep=2pt,very thick,draw,font=\ttfamily] () {cal};
 Tokenizations}

\author{%
  Brian Siyuan Zheng\uw\aspace
  Alisa Liu\uw\aspace
  Orevaoghene Ahia\uw\aspace\\
  \textbf{Jonathan Hayase}\uw\aspace
  \textbf{Yejin Choi}\stanford\aspace
  \textbf{Noah A. Smith}\uw\aiTwo\aspace\\[0.4ex]
  \uw University of Washington\aspace\aiTwo Allen Institute for AI \aspace\stanford Stanford University \\ 
  \texttt{zhengbr@cs.washington.edu}
}

\begin{document}

\maketitle

\begin{abstract}
Modern tokenizers employ deterministic algorithms to map text into a single ``canonical" token sequence, yet the same string can be encoded as many non-canonical tokenizations using the tokenizer vocabulary.
In this work, we investigate the robustness of LMs to text encoded with non-canonical tokenizations entirely unseen during training.
Surprisingly, when evaluated across 20 benchmarks, we find that instruction-tuned models retain up to 93.4\% of their original performance when given a randomly sampled tokenization, and 90.8\% with character-level tokenization. 
We see that overall stronger models tend to be more robust, and robustness diminishes as the tokenization departs farther from the canonical form. 
Motivated by these results, we then identify settings where non-canonical tokenization schemes can \textit{improve} performance, finding that character‑level segmentation improves string manipulation and code understanding tasks by up to +14\%, and right‑aligned digit grouping enhances large‑number arithmetic by +33\%.
Finally, we investigate the source of this robustness, finding that it arises in the instruction-tuning phase.
We show that while both base and post-trained models grasp the semantics of non-canonical tokenizations (perceiving them as containing misspellings), base models try to mimic the imagined mistakes and degenerate into nonsensical output, while post-trained models are committed to fluent responses.
Overall, our findings suggest that models are less tied to their tokenizer than previously believed, and demonstrate the promise of intervening on tokenization at inference time to boost performance.\footnote{Code is available at \url{https://github.com/Brianzhengca/Tokenizer-Robustness}.}
\end{abstract}

\section{Introduction}


Tokenizers segment text into a sequence of discrete tokens in the language model's (LM) vocabulary.
Most of today's LMs use deterministic subword tokenization, which produces a single canonical token sequence for a given piece of text, and further, for each whitespace-delimited word.
One commonly discussed limitation of this approach is that, by mapping byte strings to symbolic token IDs, the orthographic makeup of tokens is obscured to the LM \cite{provilkov-etal-2020-bpe, edman-etal-2024-cute}.
This can be especially harmful for LM understanding of numbers \citep{nogueira-etal-2021-investigating, thawani-etal-2021-representing, singh-etal-2024-tokenization} and morphologically rich languages \citep{arnett2025language, hofmann-etal-2021-superbizarre}, and has motivated efforts to model text directly at the byte level \citep{clark-etal-2022-canine, xue-etal-2022-byt5, wang-etal-2024-mambabyte, tay-etal-2022-charformer,yu-etal-2023-megabyte, nawrot-etal-2023-efficient, pagnoni-etal-2024-byte, ahia-etal-2024-magnet, limisiewicz-etal-2024-myte}.

\begin{figure}[t]
  \noindent\makebox[\linewidth][c]{%
    \begin{subfigure}{0.55\linewidth}
        \includegraphics[width=0.98\linewidth]{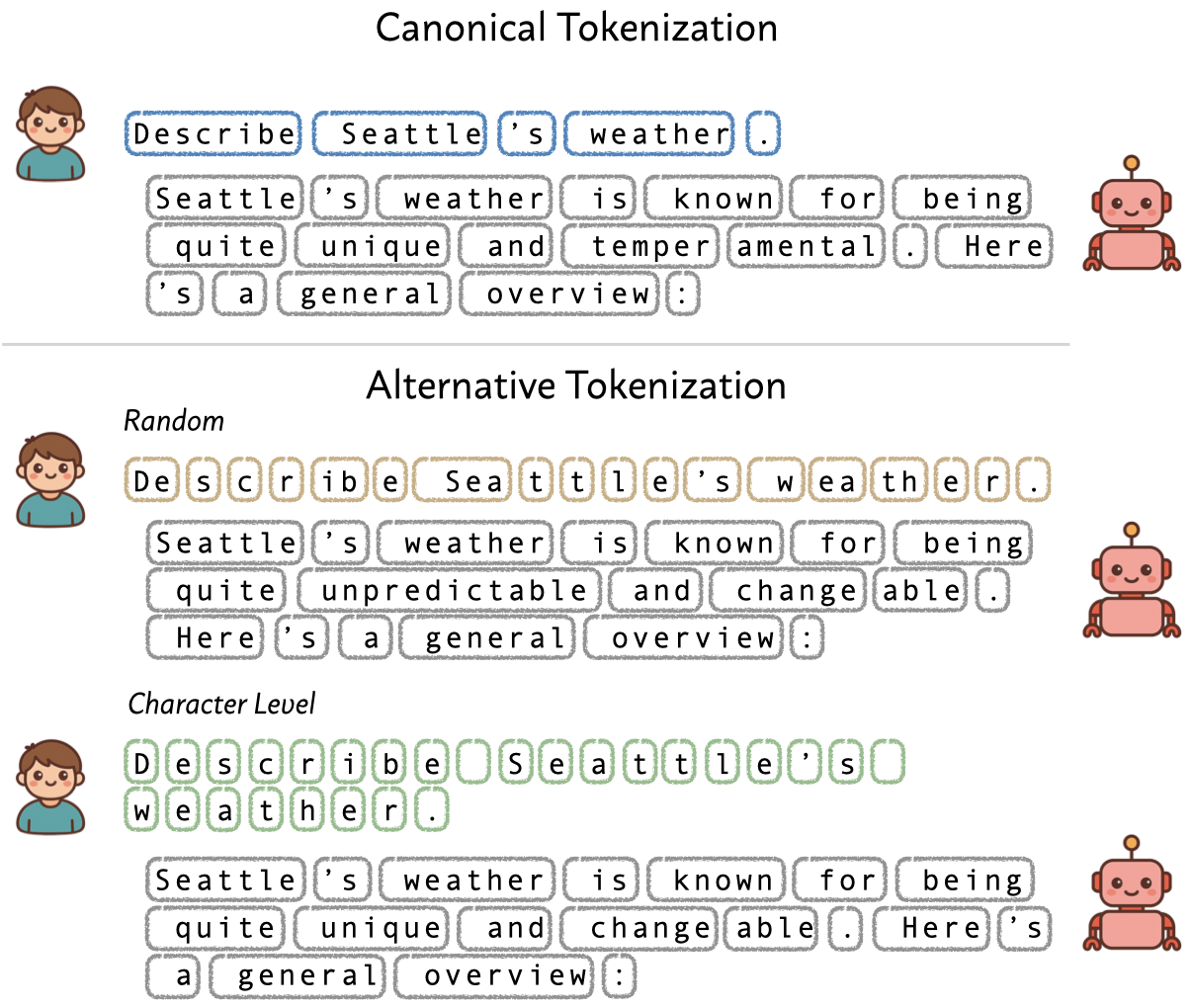}
    \end{subfigure}
    \begin{subfigure}{0.45\linewidth}
        \includegraphics[width=0.98\linewidth]{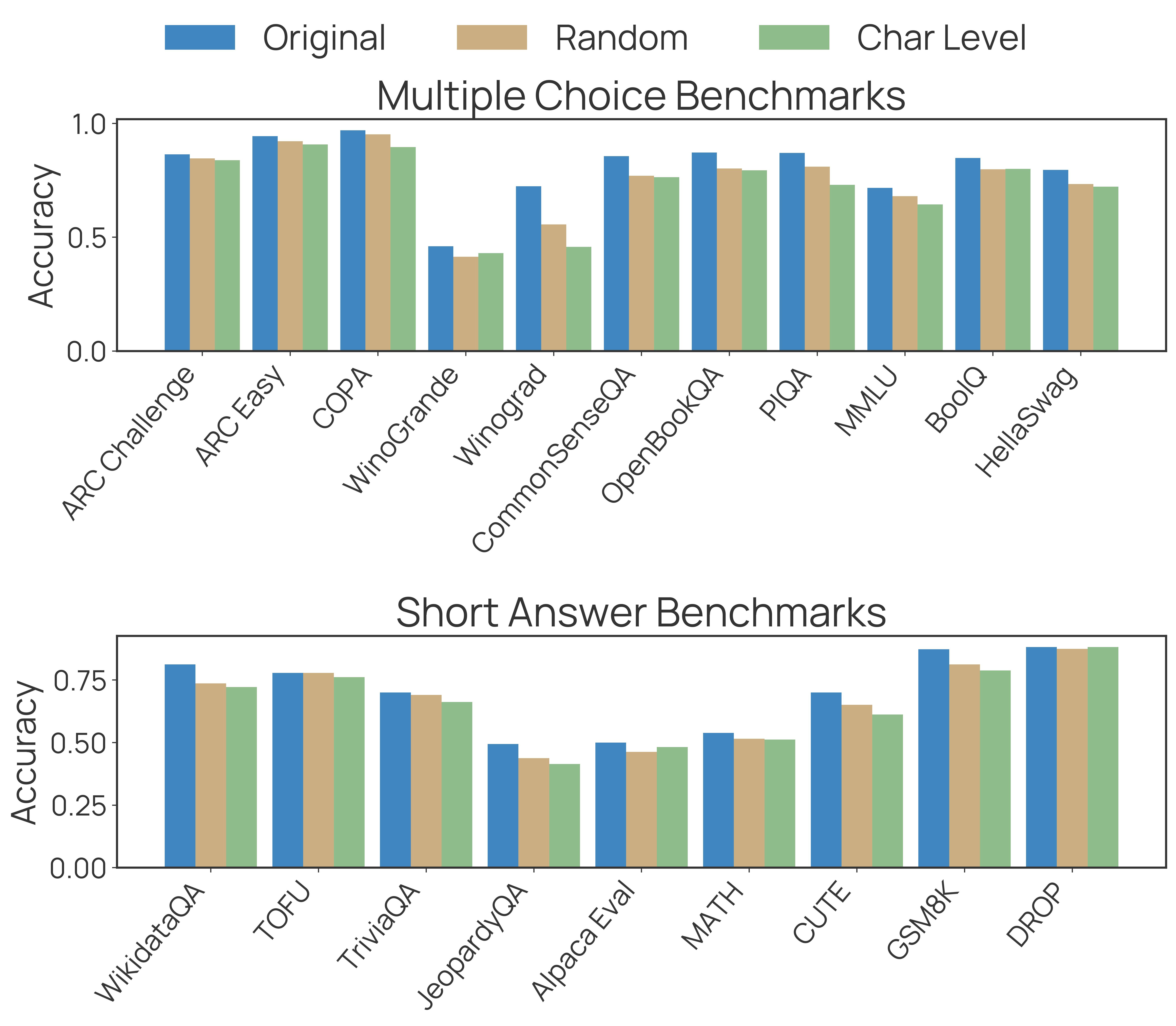}
    \end{subfigure}
  } 
  \caption{\textbf{Left:} An example of how \lm{Llama-3.1-8B-Instruct} responds when given \textcolor[HTML]{4086BF}{\textbf{canonically tokenized}} input, versus a \textcolor[HTML]{CBAE81}{\textbf{random tokenization}} and \textcolor[HTML]{8FBC8B}{\textbf{character-level tokenization}}.
  The responses are surprisingly similar, demonstrating their ability to handle non-canonical tokenizations. 
  Moreover, LMs generally respond with correctly tokenized output regardless of the tokenization scheme used for the context.
  \textbf{Right:} Performance of \lm{Qwen-2.5-7B-Instruct} across various benchmarks and tokenization schemes. Much of the original performance is preserved when given non-canonical tokenizations.
  }
  \label{fig:figure1}
  \vspace{-2em}
\end{figure}

To shed more light on this perceived limitation, in this work we study whether LMs can adapt \emph{at inference time}, without any additional training, to a different tokenization scheme than the one they were trained with.
While the tokenizer deterministically outputs a \textit{canonical tokenization} of any text into tokens (usually by applying an ordered list of merge rules), \textit{non-canonical tokenizations} of the same text using the same vocabulary are generally possible (see example in \autoref{fig:figure1}).
Here, we evaluate how LMs trained with deterministic tokenizers behave when given non-canonical tokenizations of text.
Surprisingly, we find that \textit{instruction-tuned LMs across many model families are extremely robust} to non-canonical tokenizations (\autoref{sec:experiments}).
For example, when evaluated across 20 benchmarks, \lm{Qwen-2.5-7B-Instruct} retains 93.4\% of its original performance when presented with a random non-canonical tokenization, and 90.8\% when presented with character-level tokens (see \autoref{fig:figure1}).
Thus, far from not understanding the makeup of their tokens, LMs are able to compose token sequences in entirely new ways at inference time \cite{kaplan-etal-2025-from}. 

This leads to an intriguing possibility: if LMs can process non-canonical tokenizations, can we use different tokenization schemes at inference time to improve performance?
For instance, prior work has found that better segmentation of large numbers can improve accuracy on arithmetic \citep{singh-etal-2024-tokenization, sathe-etal-2025-improving}.
Indeed, we identify several settings where non-canonical tokenization schemes improve performance for \lm{Llama-3.1-Instruct} (\autoref{sec:better_tokenizations}): character-level tokenization brings up to +14\% improvement on string manipulation and code understanding tasks, perhaps by granting LMs more direct access to orthographic cues.
Meanwhile, right-aligned digit groups, which provide a consistent grouping of digits by powers of a thousand, improves arithmetic on large numbers by +33\%.
These performance gains are achieved without any model finetuning, pointing to the promise of tokenization as a means of inference-time control.

Finally, we investigate the origins of model robustness to non-canonical tokenizations (\autoref{sec:analysis}).
Across multiple model families, we find that \textit{pretrained-only} LMs  consistently fail to produce fluent continuations given non-canonically tokenized context.
By studying models at different stages of post-training, we identify that robustness arises during the supervised instruction-tuning (SFT) phase (\autoref{subsec:when}).
We then ablate differences between pretraining and SFT procedures and find that the separation of the instruction and response as distinct turns of conversation is key (\autoref{subsec:ablations}).
From here, we provide evidence for a plausible explanation: while both base and post-trained models grasp the semantics of non-canonical tokenizations, they also perceive them as containing misspellings (\autoref{subsec:ablations}).
Base models attempt to mimic the imagined mistakes and degenerate into nonsense, whereas post-trained models are not bound by the style of the instruction and thus able to produce fluent responses.

Overall, despite being trained with deterministic tokenization, instruction-tuned LMs readily accommodate new tokenizations at inference time, suggesting that LMs are less constrained by their tokenizer than previously believed \citep{minixhofer2024zeroshot}. 
Moreover, in settings where different representations of text are beneficial, we can intervene on tokenization at inference time for performance gains.
We hope our work sheds new light on the discussion of strengths and limitations of tokenization, and points to the possibility of dynamically finding the optimal representation of text after pretraining.

\begin{table}[t]
    \centering
    \caption{
    \textbf{Evaluated across many benchmarks, models are surprisingly robust to non-canonical tokenizations of the context.}
    We show the absolute drop in performance when given a randomly sampled non-canonical tokenization (\textbf{Rand $\Delta$}) and character-level tokenization (\textbf{Char $\Delta$}), relative to the canonical (\textbf{Canon}) tokenization. We also summarize the model's ability to retain performance across benchmarks and tokenization strategies (bottom). 
    }
    \vspace{1em}
    \small
    \resizebox{\columnwidth}{!}{
    \begin{tabular}{lrrrrrrrrr}
    \toprule
        & \multicolumn{3}{c}{\textsc{Qwen-2.5-7B-Instruct}} & \multicolumn{3}{c}{\textsc{Llama-3.1-8B-Instruct}} & \multicolumn{3}{c}{\textsc{Olmo-2-7B-Instruct}}\\
        Benchmark & Canon & Rand $\Delta$ & Char $\Delta$ & Canon & Rand $\Delta$ & Char $\Delta$ & Canon & Rand $\Delta$ & Char $\Delta$\\
        \midrule
        \multicolumn{10}{c}{\textit{Multiple choice (MC)}}\\
        \midrule
        ARC-C & 86.4 & \cellcolor{red!4!white}--1.80 & \cellcolor{red!5!white}--2.60 & 76.2 & \cellcolor{red!29!white}--14.10 & \cellcolor{red!46!white}--22.40 & 77.0 & \cellcolor{red!76!white}--37.40 & \cellcolor{red!91!white}--44.60\\
        ARC-E & 94.4 & \cellcolor{red!4!white}--2.20 & \cellcolor{red!7!white}--3.60 & 91.3 & \cellcolor{red!26!white}--12.60 & \cellcolor{red!44!white}--21.50 & 85.4 & \cellcolor{red!76!white}--37.00 & \cellcolor{red!100!white}--49.00\\
        COPA & 97.0 & \cellcolor{red!4!white}--1.80 & \cellcolor{red!15!white}--7.40 & 97.2 & \cellcolor{red!20!white}--9.60 & \cellcolor{red!30!white}--14.80 & 93.8 & \cellcolor{red!44!white}--21.40 & \cellcolor{red!69!white}--33.60\\
        Winogrande & 46.0 & \cellcolor{red!9!white}--4.60 & \cellcolor{red!6!white}--3.00 & 59.6 & \cellcolor{green!4!white}+2.00 & \cellcolor{red!10!white}--5.00 & 58.6 & \cellcolor{red!16!white}--7.80 & \cellcolor{red!14!white}--7.00\\
        Winograd & 72.4 & \cellcolor{red!34!white}--16.80 & \cellcolor{red!54!white}--26.60 & 74.4 & \cellcolor{red!19!white}--9.40 & \cellcolor{red!27!white}--13.00 & 72.4 & \cellcolor{red!18!white}--8.60 & \cellcolor{red!39!white}--19.00\\
        CSQA & 85.6 & \cellcolor{red!18!white}--8.60 & \cellcolor{red!19!white}--9.20 & 77.6 & \cellcolor{red!23!white}--11.40 & \cellcolor{red!41!white}--20.00 & 75.4 & \cellcolor{red!64!white}--31.60 & \cellcolor{red!82!white}--40.00\\
        OpenbookQA & 87.2 & \cellcolor{red!14!white}--7.00 & \cellcolor{red!16!white}--7.80 & 82.0 & \cellcolor{red!28!white}--13.80 & \cellcolor{red!41!white}--20.20 & 76.2 & \cellcolor{red!63!white}--30.80 & \cellcolor{red!83!white}--40.80\\
        PIQA & 87.0 & \cellcolor{red!12!white}--6.00 & \cellcolor{red!29!white}--14.00 & 84.0 & \cellcolor{red!26!white}--12.60 & \cellcolor{red!38!white}--18.40 & 78.2 & \cellcolor{red!36!white}--17.40 & \cellcolor{red!51!white}--25.00\\
        MMLU & 71.7 & \cellcolor{red!8!white}--3.70 & \cellcolor{red!15!white}--7.30 & 68.2 & \cellcolor{red!24!white}--11.60 & \cellcolor{red!49!white}--24.00 & 59.5 & \cellcolor{red!33!white}--16.30 & \cellcolor{red!59!white}--29.10\\
        BoolQ & 84.8 & \cellcolor{red!10!white}--5.00 & \cellcolor{red!10!white}--4.80 & 86.2 & \cellcolor{red!39!white}--19.20 & \cellcolor{red!35!white}--17.20 & 71.0 & \cellcolor{red!8!white}--4.00 & \cellcolor{red!19!white}--9.20\\
        HellaSwag & 79.6 & \cellcolor{red!13!white}--6.20 & \cellcolor{red!15!white}--7.40 & 68.6 & \cellcolor{red!29!white}--14.20 & \cellcolor{red!48!white}--23.40 & 68.0 & \cellcolor{red!55!white}--26.80 & \cellcolor{red!81!white}--39.80\\
        \midrule
        \multicolumn{10}{c}{\textit{Short answer (SA)}}\\
        \midrule
        WikidataQA & 81.2 & \cellcolor{red!16!white}--7.60 & \cellcolor{red!18!white}--9.00 & 78.6 & \cellcolor{red!25!white}--12.40 & \cellcolor{red!37!white}--18.00 & 73.2 & \cellcolor{red!59!white}--28.80 & \cellcolor{red!66!white}--32.20\\
        TOFU & 77.8 & \cellcolor{green!0!white}+0.00 & \cellcolor{red!3!white}--1.70 & 82.1 & \cellcolor{green!3!white}+1.70 & \cellcolor{green!2!white}+0.80 & 82.9 & \cellcolor{red!26!white}--12.80 & \cellcolor{red!49!white}--23.90\\
        TriviaQA & 70.0 & \cellcolor{red!2!white}--1.00 & \cellcolor{red!8!white}--3.80 & 76.6 & \cellcolor{red!20!white}--9.80 & \cellcolor{red!28!white}--13.60 & 70.0 & \cellcolor{red!45!white}--22.20 & \cellcolor{red!71!white}--34.80\\
        JeopardyQA & 49.4 & \cellcolor{red!11!white}--5.60 & \cellcolor{red!16!white}--8.00 & 43.6 & \cellcolor{red!4!white}--2.20 & \cellcolor{red!21!white}--10.20 & 42.6 & \cellcolor{red!44!white}--21.60 & \cellcolor{red!49!white}--24.20\\
        AlpacaEval & 50.0 & \cellcolor{red!8!white}--3.70 & \cellcolor{red!4!white}--1.80 & 50.0 & \cellcolor{red!12!white}--5.70 & \cellcolor{red!15!white}--7.50 & 50.0 & \cellcolor{red!4!white}--2.10 & \cellcolor{red!23!white}--11.30\\
        MATH & 53.9 & \cellcolor{red!5!white}--2.40 & \cellcolor{red!6!white}--2.70 & 32.0 & \cellcolor{red!9!white}--4.20 & \cellcolor{red!20!white}--9.70 & 22.7 & \cellcolor{red!11!white}--5.20 & \cellcolor{red!19!white}--9.20\\
        CUTE & 70.0 & \cellcolor{red!10!white}--4.90 & \cellcolor{red!18!white}--8.80 & 68.0 & \cellcolor{red!23!white}--11.10 & \cellcolor{red!31!white}--15.30 & 55.3 & \cellcolor{red!21!white}--10.20 & \cellcolor{red!12!white}--5.70\\
        GSM8K & 87.3 & \cellcolor{red!12!white}--6.10 & \cellcolor{red!17!white}--8.50 & 82.0 & \cellcolor{red!24!white}--11.70 & \cellcolor{red!33!white}--16.00 & 73.9 & \cellcolor{red!47!white}--23.10 & \cellcolor{red!73!white}--35.80\\
        DROP & 88.2 & \cellcolor{red!2!white}--0.80 & \cellcolor{green!0!white}+0.00 & 88.8 & \cellcolor{red!1!white}--0.60 & \cellcolor{red!10!white}--5.00 & 77.0 & \cellcolor{red!11!white}--5.60 & \cellcolor{red!16!white}--7.60\\
        \midrule
        \multicolumn{10}{c}{\textit{Overall}}\\
        \midrule
        \multicolumn{2}{l}{Avg MC Retention (\%)} & $92.4_{\pm 5.97}$ & $89.2_{\pm 9.33}$ && $85.6_{\pm 6.86}$ & $76.8_{\pm 7.99}$ && $71.2_{\pm 14.7}$ & $59.2_{\pm 17.8}$\\
        \multicolumn{2}{l}{Avg SA Retention (\%)} & $94.6_{\pm 3.97}$ & $92.7_{\pm 5.35}$ && $90.3_{\pm 6.78}$ & $82.7_{\pm 9.65}$ && $75.4_{\pm 15.1}$ & $65.4_{\pm 17.4}$\\
        \multicolumn{2}{l}{Avg Overall Retention (\%)} & $93.4_{\pm 5.15}$ & $90.8_{\pm 7.81}$ && $87.7_{\pm 7.05}$ & $79.4_{\pm 9.05}$ &&$73.1_{\pm 14.7}$ & $62.0_{\pm 17.5}$\\
    \bottomrule
    \end{tabular}}
    \label{tab:drop_stats}
\end{table}

\section{Language Models are Robust to Non-Canonical Tokenizations}\label{sec:experiments}
In our main experiments, we evaluate the robustness of LMs to non-canonical tokenizations by comparing their performance on downstream tasks when given different tokenizations of the input. 

\subsection{Background}\label{subsec:background}
Most LMs today, and all the models we study, use the \textit{Byte-Pair Encoding} (BPE) \cite{sennrich-etal-2016-neural} algorithm for tokenization.
The BPE tokenizer is learned by splitting a corpus of text into bytes, which form the initial vocabulary, then iteratively merging the most frequent pair of tokens into a new token that is added to the vocabulary.
To encode a new text, it is  split into bytes, and the learned merges are applied in the same order.
As a result, a BPE tokenizer always produces the same token sequence for the same text.
Further, because BPE tokens do not cross whitespace boundaries, the same whitespace-delimited word is always represented with the same token or token sequence.

A natural observation is that given a tokenizer vocabulary, there exist many token sequences that decode to the same text.
For instance, \whitespace \textit{cat} could be tokenized as [\whitespace\texttt{cat}], [\whitespace, \texttt{cat}], [\whitespace, \texttt{c}, \texttt{at}], [\whitespace, \texttt{c}, \texttt{a}, \texttt{t}], etc.
In general, the number of non-canonical tokenizations grows exponentially with the length of the text.
Many previous works have argued that the probability of a string should be calculated as the sum of probabilities of all possible tokenizations \citep{cao-rimell-2021-evaluate, chirkova-etal-2023-marginalize, geh-etal-2024-signal}.
However, less attention has been paid to how non-canonical tokenizations affect LMs in generative settings. 

\subsection{Setup}
We consider two non-canonical tokenization schemes. 
(1) \textit{Random tokenization} produces a tokenization (uniformly at random) from the set of tokenizations more granular than the canonical one.
This can be achieved by recursively splitting individual tokens into a valid pair of tokens, similarly to \citep{sims-etal-2025-stochastok}; the pseudocode and a proof of correctness is provided in \autoref{sec:appendix_random_segmentation}. 
(2) \textit{Character-level tokenization} decomposes the string into character tokens, i.e., using no subword token from the vocabulary.
For text containing only English letters and punctuation (where each character is exactly one byte), this produces the most granular possible tokenization.

\begin{figure}
    \centering
    \includegraphics[width=\textwidth]{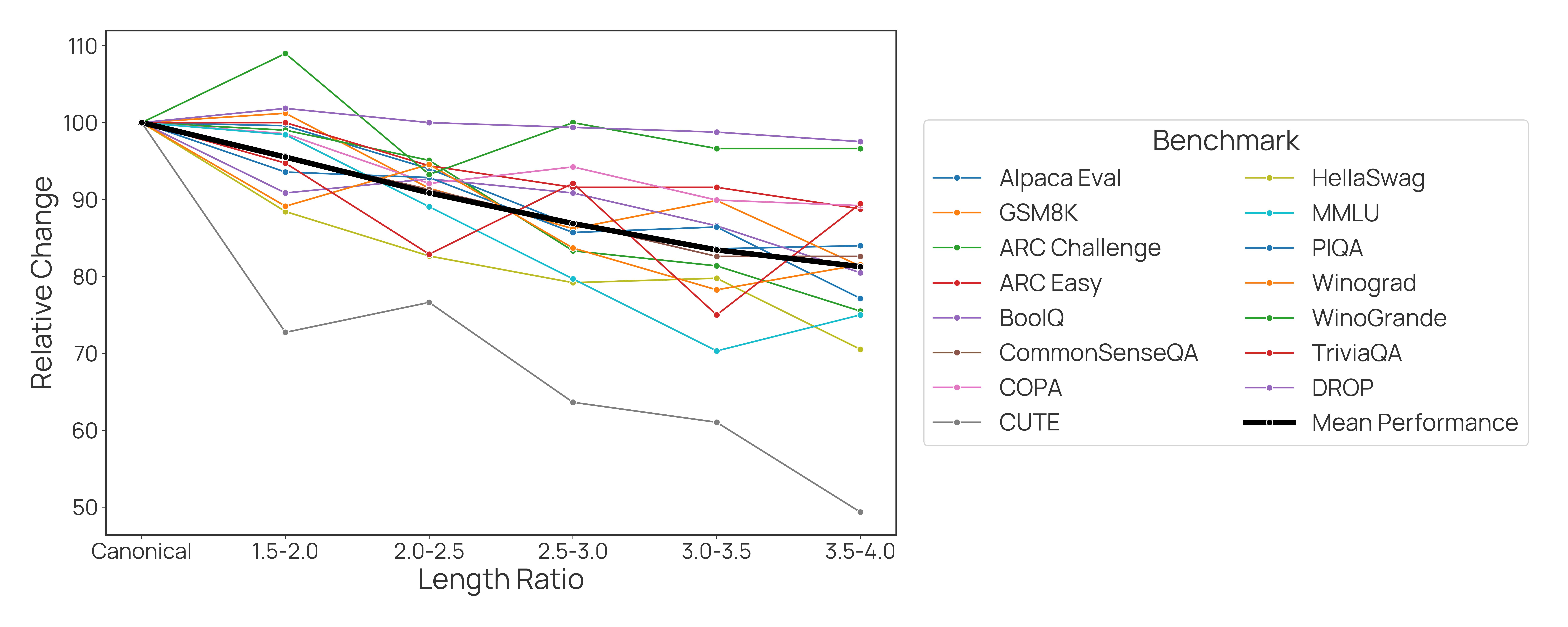}
    \caption{\textbf{Model performance generally declines as the tokenization becomes more granular.} We achieve variation in tokenization length using different values of $p$ in BPE-dropout, and group tokenizations into buckets based on how many times longer it is than the canonical tokenization.
    }
    \label{fig:granularity}
\end{figure}

We consider three models, \lm{Llama-3.1-8B-Instruct}~\citep{grattafiori2024llama3herdmodels}, \lm{OLMO2-7B-Instruct}~\citep{olmo-etal-2024-2olmo2furious}, and \lm{Qwen-2.5-7B-Instruct}~\citep{qwen2025qwen25technicalreport}, which we evaluate on 20 benchmarks shown in \autoref{tab:drop_stats}.
Please see \autoref{gen_bchmk_eval_dtls} for further description of the datasets and evaluation setup.




\subsection{Results}
Shown in \autoref{tab:drop_stats}, while random tokenization consistently leads to worse performance compared to the canonical tokenization, the effect is small.
On average across benchmarks, \lm{Qwen-2.5} retains 93.4\% of its performance when given random tokenization, followed by \lm{Llama-3.1} at 87.7\% and \lm{Olmo-2} at 73.1\%.
The performance drops further with character-level tokenization, with retention of 90.8\%, 79.4\%, and 62.0\% for the three models, respectively.
This ranking of models in terms of retention is consistent with their ranking in absolute accuracy (under canonical tokenization), suggesting that stronger models are generally more robust to non-canonical tokenization strategies. 

We also observe that all models retain performance better on short answer (SA) benchmarks (where the model generates an output in free-form) compared to multiple choice (MC) benchmarks (where the model is instructed to directly output the correct answer choice).
In addition, LMs consistently \emph{produce} correct token sequences even when conditioning on non-canonical tokenizations. We hypothesize that, in the SA setting, models benefit from eventually conditioning on recent correctly-tokenized context.


\subsection{Analysis: How does granularity of the tokenization affect robustness?}\label{subsec:fineness}

We next study whether tokenization fine-grainedness correlates in general with model robustness.
We measure the fine-grainedness of a given non-canonical tokenization by how many times longer it is (in tokens) than the canonical tokenization, which we call the ``length ratio.''
Finer-grained tokenizations have higher ratios, while coarser ones have ratios closer to 1.
We produce tokenizations with diverse length ratios by applying BPE dropout \citep{provilkov-etal-2020-bpe} with $p\in[0.1,0.2,...,0.9]$, which controls the probability with which each merge is dropped. 
(High $p$ leads to finer-grained segmentations, and $p=0.0$ corresponds to conventional BPE.)


\autoref{fig:granularity} shows the relationship between the length ratio and the average performance retention relative to canonical tokenization, with finer-grained tokenization generally leading to worse performance.
When performance retention is averaged across tasks, the negative correlation is statistically significant under Kendall's $\tau$ with \(p = 0.003\).

\section{Can non-canonical tokenizations \textit{improve} model performance?}\label{sec:better_tokenizations}

\begin{table}[t!]
    \centering\small
    \caption{\textbf{Examples from tasks} we construct where non-canonical tokenizations lead to improved performance for \lm{Llama-3.1-7B-Instruct} (\autoref{sec:better_tokenizations}).}
    \vspace{1em}
    \begin{tabular}{p{390pt}}
    \toprule
        \textbf{Counting characters:} \texttt{Count the number of the letter \textquotesingle r\textquotesingle\, in the word strawberry.} \\\midrule
        \textbf{Acronyms:} \texttt{Come up with a sequence of words where the first letters would form this acronym: isman} \\\midrule
        \textbf{Codeline Description:} 
        \texttt{What does the following code do:} \\
        \texttt{\colorbox{gray}{\{code block\}}}\\
        \texttt{A. Counts paths from a point to reach Origin}\\
        \texttt{B. Program to check if a matrix is symmetric }\\
        \texttt{C. Longest subsequence from an array of pairs having first element increasing and second element decreasing .}\\
        \texttt{D. Count the number of strings in an array whose distinct characters are less than equal to M}\\\midrule
        \textbf{Arithmetic:} \texttt{8492079913 + 4877278482 =} \\
        \bottomrule
    \end{tabular}
    \label{tab:example_prompts}
    \vspace{-1em}
\end{table}


If LMs can process non-canonical tokenizations, this points to the exciting possibility that tokenization schemes can be modified completely at inference-time.
This would be useful if, in certain settings, there exists a better representation of text than what the tokenizer produces.
In this section, we develop a suite of tasks that intuitively require understanding of the orthography of the text, and show that \lm{Llama-3.1-8B-Instruct} performs better under non-canonical tokenization schemes.

\subsection{Tasks}
Please see \autoref{tab:example_prompts} for an example question in each task and \autoref{subsec:perf_improve} for further details on dataset construction.
For all tasks except Arithmetic, we use character-level tokenization.

\paragraph{Counting Characters}
This task asks the model to count the number of occurrences of the most common letter in 5-10 character tokens in \lm{Llama-3.1}'s vocabulary, and contains 1001 samples. 

\paragraph{Acronyms}
This task asks models to generate a list of words whose first letters form a given acronym.
We construct 3594 5-letter acronyms by sampling each letter uniformly at random from the alphabet. 

\paragraph{Code Description}
For a more real-world application, we construct a task where the model is given a code snippet and asked to identify the function of the code in natural language from four MC options.
The setup is inspired by the Codeline Description task from BIG-Bench \citep{srivastava2023beyond}, but to increase the difficulty we use more complex code snippets and corresponding natural language descriptions from XLCoST \cite{zhu2022xlcost}. 
To collect incorrect answers, we sample three other code descriptions from the dataset. This task contains 4800 samples across 6 programming languages.

\paragraph{Arithmetic}
Prior work has suggested that arithmetic is difficult for LMs in part due to poor segmentation of digits \citep{nogueira-etal-2021-investigating, thawani-etal-2021-representing}.
We curate a simple arithmetic dataset by constructing addition and subtraction tasks for 10 digit numbers. 
Here, we use a different segmentation strategy.
The \lm{Llama-3.1} tokenizer segments numbers into groups of three left-to-right (e.g., \textit{1000000} is encoded as [``\texttt{100}", ``\texttt{000}", ``\texttt{0}"]), due to the pretokenization regular expression looking for matches greedily from the left.
Inspired by \citep{singh-etal-2024-tokenization}, we instead segment digits into groups of three right-to-left (e.g., [``\texttt{1}", ``\texttt{000}", ``\texttt{000}"]). 
This task contains 1000 addition and subtraction questions in total. 


\begin{table}[t]
    \centering
    \caption{\textbf{On several tasks, \lm{Llama-3.1-8B-Instruct} achieves \textit{better} performance when using a non-canonical tokenization scheme}. For the first four tasks, the input is tokenized at the character level; for Arithmetic, we segment digits into groups of three digits from right to left (instead of the usual left to right). On all tasks, we observe a large performance improvement from using the alternative tokenization scheme.
    }
    \vspace{1em}
    \begin{tabular}{lccc}
    \toprule
    Task & Canonical & Alternative & $\Delta$ \\
    \midrule
         Counting Characters & 66.5 & \textbf{73.5} & +6.99\\
         Acronyms & 49.7 & \textbf{57.4} & +7.74 \\
         Code Description & 68.6 & \textbf{82.9} & +14.3\\
         Arithmetic & 36.5 & \textbf{70.2} & +33.7\\
    \bottomrule
    \end{tabular}
    \label{tab:task_improvements}
\end{table}

\subsection{Results}
Shown in \autoref{tab:task_improvements}, in all the tasks we construct, the non-canonical tokenization strategy leads to substantially better performance compared to the canonical tokenization.
In particular, we observe a +14.3\% improvement on code description and +33.7\% on arithmetic.
Our results show that the tokenization scheme used in training is not necessarily the optimal one at inference-time, and replacing them with intuitively meaningful tokenizations can bring substantial performance gains.
We leave automatically identifying the optimal tokenization as a promising direction for future work.

\section{Investigating the Source of Robustness}\label{sec:analysis}

Thus far, our experiments have used post-trained ``instruct'' models.
In this section, we find that pretrained-only models are actually unable to produce fluent continuations of unusually tokenized context (\autoref{subsec:when}), perform ablations to identify the conditions enabling robustness (\autoref{subsec:ablations}), and finally provide support for an explanation of why generative robustness arises during post-training (\autoref{subsec:understanding}).

\subsection{When does robustness appear in model training?}\label{subsec:when}

We first quantify the robustness of models at different stages of the model development pipeline by using the \lm{Olmo2} and \lm{Tulu3} \citep{lambert-etal-2025-tulu3} model families which include the base, SFT, DPO, and final instruct models. 
For simplicity, we focus on AlpacaEval and use character-level tokenization. 
For base models, we construct the prompt by placing the instruction in a question-answer template (\texttt{Question:\,\colorbox{gray}{\{instruction\}}\textbackslash nAnswer:}).
We define three simple measures of generation quality.

\paragraph{Spelling} We measure the proportion of (whitespace-delimited) words in the generation that can be found in a collection of the top 10,000 most common English words.\footnote{\url{https://github.com/first20hours/google-10000-english}}

\paragraph{Grammaticality} 
We use LanguageTool's grammar checker\footnote{\url{https://github.com/languagetool-org/languagetool}} to count the number of grammatical mistakes, which we normalize by the number of words in the generation and subtract from 1 to produce a grammaticality score where higher is better.


\paragraph{Win rate}
To measure overall generation quality, we use \texttt{alpaca\textunderscore eval\textunderscore gpt4} as an LM judge in the AlpacaEval framework and report the win rate of the generation given alternative against canonical tokenizations of the context. 
Unlike the previous two metrics, this measures not only the quality of the generation but also its relevance to the context.

\begin{figure}[t]
  \noindent\makebox[\linewidth][c]{%
    \includegraphics[width=0.8\linewidth]{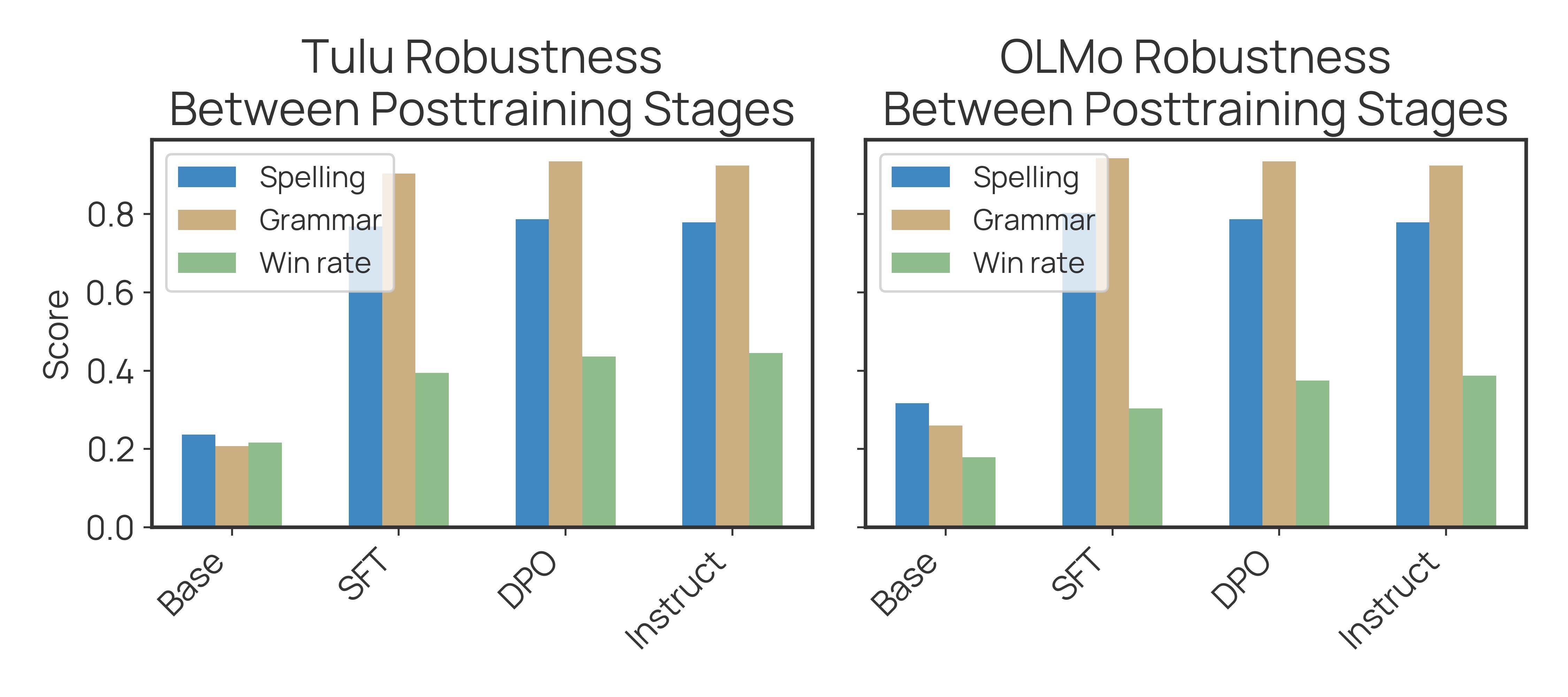}
  }
  \caption{\textbf{Pretrained-only models completely fail to generate coherent output conditioned on non-canonical tokenizations of context; robustness is gained in the SFT stage.} We evaluate the spelling, grammaticality, and AlpacaEval win rate of model generations.
  Note that since \lm{Tulu3} uses \lm{Llama-3} as the base model, its base scores are computed using \lm{Llama-3}'s base model scores.  
  }
  \label{fig:when-robustness}
\end{figure}

Shown in \autoref{fig:when-robustness}, the base models of \lm{Olmo2} and \lm{Llama-3.1} are both unable to produce sensible output conditioned on character-level tokenizations of context, scoring at best 0.317 on spelling and 0.260 on grammaticality.
Qualitatively, generations are extremely difficult to parse and often involve odd character substitutions and repetitions (e.g., \texttt{Yoou}, \texttt{haviin}).
Despite this, they sometimes reflect an understanding of the prompt.
Consider, for example,

\begin{table}[h]
    \centering
    \small
    \begin{tabular}{l}
        \texttt{\makecell*[{{p{350pt}}}]{Question: I like to host guests at my home from time to time [...] Can you give me a recipe for Canjeero?\\
        Answer: \colorbox[HTML]{d6eaf8}{I aam glade tio hear tio hear tio hear tio hear that yoou enjoy}\\
        \colorbox[HTML]{d6eaf8}{haviin  gauests at  yoour  hoome  an  tio keeep  tio keeep  tio keeep}}}\\
    \end{tabular}
\end{table}

In contrast, the post-trained models are more robust across all three metrics, with \textit{much of the improvement coming from the SFT stage alone}. 

\subsection{Why do instruction-tuned models become robust?}\label{subsec:ablations}

We first replicate the finding from \autoref{subsec:when} that SFT yields robustness to non-canonical tokenizations by finetuning the \lm{Llama-3.2-1B} base model on the \lm{Tulu 3 SFT Personas Instruction Following} dataset.
Then, we perform the following interventions on the SFT training data and procedure to shed light on the possible source.

\paragraph{Gradient over full sequence}
SFT on instruction-response pairs conventionally uses a loss mask over the instruction tokens, so that only the response tokens contribute to the loss. 
We remove this loss mask and instead compute gradients over the entire instruction and response. 

\paragraph{Question/answer template}
We replace the chat template with a simple question-answer template, \texttt{Question:\,\colorbox{gray}{\{instruction\}} Answer:\,\colorbox{gray}{\{response\}}}, both for training and evaluation.

\paragraph{Removing the chat template}
We remove the chat template by concatenating the instruction and response without any special formatting.
In evaluation, we again provide the instruction alone.

\paragraph{Removing the instruction}
After SFT training, the LM's goal is no longer to continue a given text prefix, but rather to generate a response to the given instruction.
To ablate the nature of the data itself, we take only the responses from the SFT data, and randomly split each into a new ``prompt'' and ``response,'' which we format with the SFT template.\footnote{We match the instruction length distribution by counting the number of tokens $n$ in the original instruction, and formatting the first $n$ tokens of the response as the new ``instruction.''}
At test time, we similarly provide an incomplete response within ``instruction'' tags.
Since the purpose of the passage is generally inferrable from the first few words of the gold response (``\textit{Sure, here's a recipe for Kubdari...}''), we are able to evaluate generated responses under the same AlpacaEval framework.

\begin{figure}[t]
  \centering
  \includegraphics[width=0.7\textwidth]{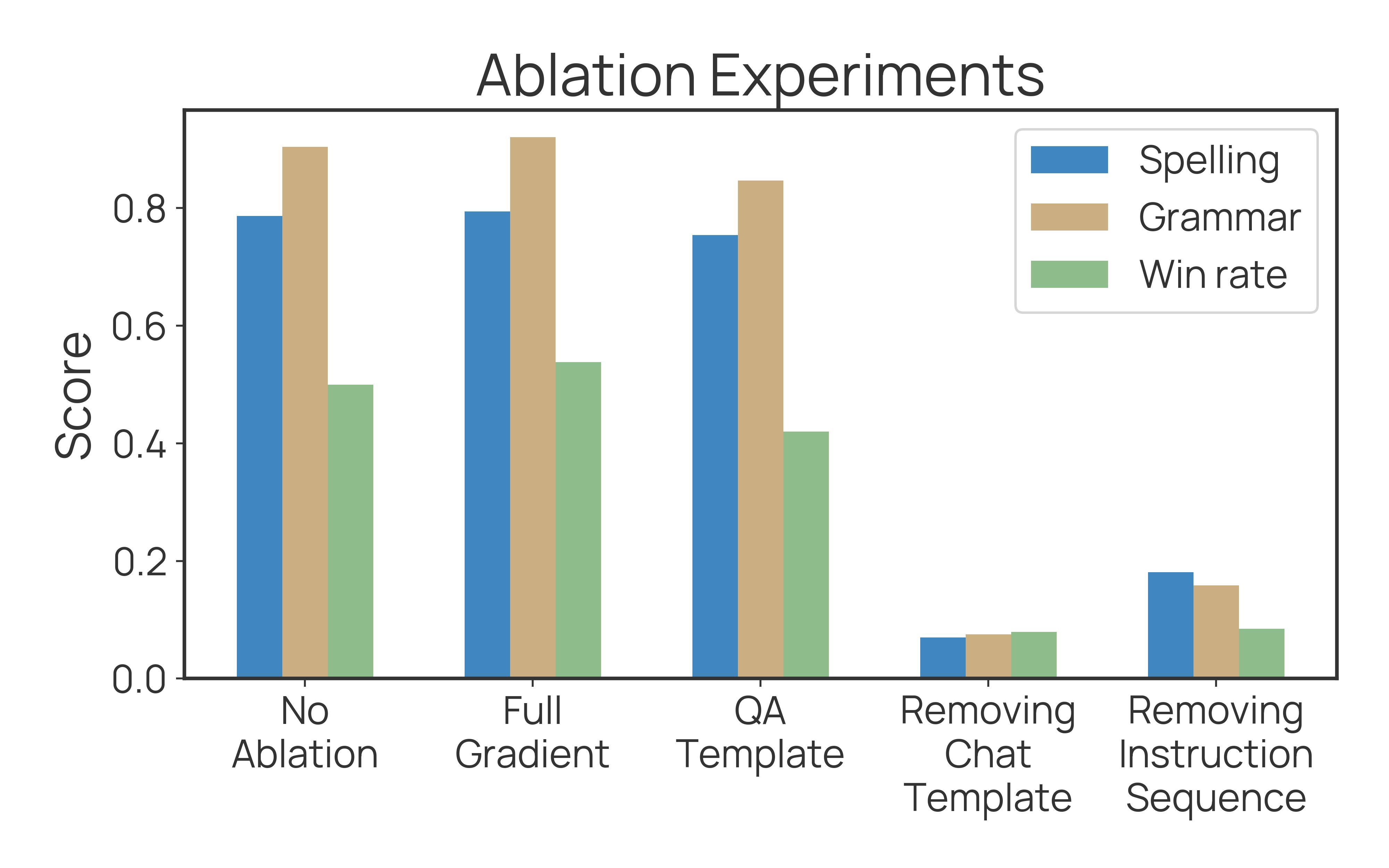}
  \caption{Ablations on the SFT training data and procedure indicate that the separation of the context and expected continuation --- as different turns of dialogue demarcated with a special token --- is key to robustness to non-canonical tokenizations.}
  \label{fig:ablation}
\end{figure}

Our results are shown in \autoref{fig:ablation}.
We replicate the finding that SFT (\textbf{No ablation}) leads the model to be able to handle non-canonical tokenizations.
This persists when computing gradients over the entire instruction and response (\textbf{Full gradient}) so that the training procedure matches regular pretraining. 
Replacing the original chat template with a simple question-answer template (\textbf{QA template}) also maintains model robustness. 
However, the usage of a template is crucial --- when directly concatenating the instruction and response (\textbf{Removing chat template}), the model fails to produce coherent generations, with the spelling score dropping from 0.786 in the no ablation setting to 0.0698.
Inserting the chat template into pretraining-style data (\textbf{Removing the instruction}) also does not yield robustness, with a spelling and grammaticality scores remaining low at 0.181 and 0.158, respectively.
Overall, these findings suggest that in order for the LM to generate fluent continuations given non-canonical tokenizations, the context and expected continuation need to represent separate turns of dialogue, and additionally, be demarcated with a special template.

\subsection{Disentangling understanding from generation}\label{subsec:understanding}

One plausible explanation for our findings thus far is that both base and instruction-tuned models grasp the semantics of non-canonical tokenizations, yet falsely perceive them as containing misspellings.
While base LMs attempt to faithfully continue these mistakes and degenerate into nonsensical output, instruction-tuned models are trained to provide fluent responses regardless of the instruction, leading to the results observed in \autoref{subsec:when}.
To test this hypothesis, we construct two simple tests:

\begin{enumerate}
    \item \textbf{Word Repeat}: To determine if a model perceives the meaning of a word with non-canonical tokenization, we prompt the model to repeat a given word (while correcting any typos).
    \item \textbf{Identifying Misspellings}: To determine if a model perceives a misspelling, we ask it to identify the word with a misspelling among two options: a (correctly tokenized) misspelling of a word and an non-canonical tokenization of that word (correctly spelled).
\end{enumerate}

\begin{table}
    \centering
    \caption{Both base and instruct models from the \lm{Llama-3.1-8B} family recognize words represented with non-canonical tokenizations (performing well on \textbf{Word Repeat}), but incorrectly perceive that there are misspellings (performing at random on \textbf{Identifying Misspellings}).}
    \vspace{1em}
    \begin{tabular}{lrr}
    \toprule
        & Word Repeat & Identifying Misspellings \\\midrule
        Base model & 90.8 & 48.2 \\
        Instruct model & 92.0 & 55.8 \\
     \bottomrule
    \end{tabular}
    \label{tab:understanding}
\end{table}

Results are shown in \autoref{tab:understanding}.
Consistent with our hypothesis, we find that both the base and instruct models from the \lm{Llama-3.1-8B} family score highly ($>90\%$) on Word Repeat.
This means that the base model, despite its poor performance in \autoref{subsec:when}, actually recognizes the correct form of non-canonical tokenizations as well as its post-trained counterpart.
In addition, both models perform at random when asked to distinguish non-canonical tokenization from true misspellings.
In other words, the instruct model produces fluent responses (\autoref{sec:experiments}) while interpreting the instruction as heavily misspelled!
While instruct models evidently overcome this, the base model likely attempts to mimic the (perceived) idiosyncratic surface form, thus producing nonsensical (yet sometimes relevant) outputs.



\section{Related Work}

The extent to which LMs are limited by their tokenization is a topic of much debate, with the story evolving as LMs become larger and more capable.

\paragraph{Character-level understanding in tokenizer-based models}
It is commonly argued that tokenization obscures orthographic information about tokens from the LM, leading to unexpected failures \cite{edman-etal-2024-cute, chai-etal-2024-tokenization, wang-etal-2024-tokenization}.
As a result, there have been many efforts towards linguistically-informed tokenization that make derivational, compound, and morphological boundaries within words explicit \cite{klein-tsarfaty-2020-getting, hofmann-etal-2021-superbizarre, hofmann-etal-2022-embarrassingly, yehezkel-pinter-2023-incorporating, bauwens-delobelle-2024-bpe}.
Similarly, BPE-dropout \cite{provilkov-etal-2020-bpe} and related methods \cite{sims-etal-2025-stochastok} introduce variation in training in how a given string is tokenized in order to make models more robust to rare, misspelled, or unseen words.

However, there is other evidence that LMs naturally overcome these limitations.
For instance, token embeddings have been found to robustly encode character-level information, especially in larger models \cite{kaushal-mahowald-2022-tokens, itzhak-levy-2022-models}.
This may be because word variants that do not share tokens in common (consider e.g., \texttt{[\whitespace dictionary]} and \texttt{[\whitespace diction, aries]}, as tokenized by GPT-2) incentivize the model to learn spelling as a general solution to understanding their relations \cite{kaushal-mahowald-2022-tokens}.
Other works argue that LMs maintain an implicit vocabulary, and can compose arbitrary token sequences (including non-canonical ones) into useful higher-level representations \cite{feucht-etal-2024-token, kaplan-etal-2025-from}.
Even in domains like biomedical text where terms are highly agglutinative, using tokenizers that segment on meaningful components does not lead to improved models \cite{jimenez-gutierrez-etal-2023-biomedical}.
Recent works have even found that coarser \textit{superword} tokenization \citep{liu-etal-2025-superbpe, schmidt-etal-2025-boundless}, which capture common word sequences in a single token, preserve character-level understanding while providing benefits in compression and downstream performance.

Our work informs this conversation by showing that LMs can effectively leverage character-level knowledge of their tokens and glean potential benefits of improved representation at inference time.

\paragraph{Partial token problem}
A related but distinct problem is the \textit{partial token problem} (also known as \textit{tokenization bias} or the \textit{prompt boundary problem}) where the prompt ends with the prefix of a valid token, causing the model to assign unexpectedly low probability to the completion of that token.
Many works have found that this continues to compromise a serious failure mode for frontier LMs \cite{xu-etal-2026-finish, phan-etal-2025-exact, lundberg-2023-art, vieira-etal-2025-from, hayase-etal-2025-sampling}.
In particular, \lm{Deepseek v3} \citep{deepseek-2025-deepseekv3} aims to improve robustness to partial punctuation tokens by randomly splitting some proportion of multi-punctuation tokens into smaller tokens during training, though they do not present experiments with this ablation.
We note that these results are not inconsistent with ours --- together, they suggest that while models are very unlikely to \textit{generate} non-canonical tokenizations, they can nonetheless understand them in the context history.

\paragraph{Non-canonical tokenizations}
It has long been recognized that there are many possible ways to segment a string into tokens with a fixed vocabulary \cite{church-2020-emerging}, which in principle should be considered in the calculation of a string's likelihood \cite{cao-rimell-2021-evaluate, chirkova-etal-2023-marginalize, geh-etal-2024-signal}.
Previous work has briefly touched on non-canonical tokenization in the context of self-supervised evaluation \cite{jain2023bringdataselfsupervisedevaluation} and defense against adversarial attacks \cite{jain2023baselinedefensesadversarialattacks}.
In contemporaneous work, \citet{geh-etal-2025-adversarial} also show that non-canonical tokenizations can be constructed adversarially to trigger unsafe completions.
In contrast, we provide a more systematic study of LM robustness using benchmark evaluations and additionally study its source.

Somewhat relatedly, other works have provided algorithms for sampling at the character- or byte-level from tokenizer-based LMs \citep{phan-etal-2024-understanding, vieira-etal-2024-language, athiwaratkun-etal-2024-token, hayase-etal-2025-sampling}.
Together, these directions suggest that despite being trained with one deterministic tokenization scheme, LMs can both condition on and produce token sequences over a different (sub)vocabulary.

\section{Conclusion}
Despite being trained with deterministic tokenization algorithms, we show that instruction-tuned language models are surprisingly robust to token sequences not seen in training.
In certain domains, such as arithmetic or code, more intuitively meaningful tokenizations can even be swapped-in at inference time for improved performance. 
We analyze the source of this robustness, and find that while the base and instruct models both perceive the semantics of non-canonical tokenizations, only instruct models are capable of providing fluent continuations.
Our work demonstrates a way in which LMs are not necessarily tied to tokenizer they were trained with, and highlights the potential of finding more optimal representations of text after pretraining.

\subsubsection*{Acknowledgments}
We would like to thank Joseph An and Ricky Koppolu, as well as the broader UW NLP community, for helpful conversations about this work. 
AL and JH are supported by the NSF Graduate Research Fellowship.

\bibliographystyle{abbrvnat}
\bibliography{anthology,custom}

\newpage
\section*{NeurIPS Paper Checklist}

\begin{enumerate}

\item {\bf Claims}
    \item[] Question: Do the main claims made in the abstract and introduction accurately reflect the paper's contributions and scope?
    \item[] Answer: \answerYes{}
    \item[] Justification: We provide support in the paper for all claims in the abstract.

\item {\bf Limitations}
    \item[] Question: Does the paper discuss the limitations of the work performed by the authors?
    \item[] Answer: \answerYes{}
    \item[] Justification: We clearly identify questions that we leave to future work and differentiate hypotheses from claims supported by evidence.

\item {\bf Theory assumptions and proofs}
    \item[] Question: For each theoretical result, does the paper provide the full set of assumptions and a complete (and correct) proof?
    \item[] Answer: \answerYes{}
    \item[] Justification: Our proof in the appendix is complete and provides the full set of assumptions.

    \item {\bf Experimental result reproducibility}
    \item[] Question: Does the paper fully disclose all the information needed to reproduce the main experimental results of the paper to the extent that it affects the main claims and/or conclusions of the paper (regardless of whether the code and data are provided or not)?
    \item[] Answer: \answerYes{}
    \item[] Justification: This information will be provided in the appendix for the appendix deadline.

\item {\bf Open access to data and code}
    \item[] Question: Does the paper provide open access to the data and code, with sufficient instructions to faithfully reproduce the main experimental results, as described in supplemental material?
     \item[] Answer: \answerYes{}
     \item[] Justification: We plan to open-source our code shortly after submission.

 \item {\bf Experimental setting/details}
     \item[] Question: Does the paper specify all the training and test details (e.g., data splits, hyperparameters, how they were chosen, type of optimizer, etc.) necessary to understand the results?
     \item[] Answer: \answerYes{}
     \item[] Justification: This information will be provided in the appendix.

 \item {\bf Experiment statistical significance}
     \item[] Question: Does the paper report error bars suitably and correctly defined or other appropriate information about the statistical significance of the experiments?
     \item[] Answer: \answerYes{}
     \item[] Justification: We provide information about standard deviation and statistical significance.
     \item[] Guidelines:
     \begin{itemize}
         \item The answer NA means that the paper does not include experiments.
         \item The authors should answer "Yes" if the results are accompanied by error bars, confidence intervals, or statistical significance tests, at least for the experiments that support the main claims of the paper.
         \item The factors of variability that the error bars are capturing should be clearly stated (for example, train/test split, initialization, random drawing of some parameter, or overall run with given experimental conditions).
         \item The method for calculating the error bars should be explained (closed form formula, call to a library function, bootstrap, etc.)
         \item The assumptions made should be given (e.g., Normally distributed errors).
         \item It should be clear whether the error bar is the standard deviation or the standard error of the mean.
         \item It is OK to report 1-sigma error bars, but one should state it. The authors should preferably report a 2-sigma error bar than state that they have a 96\ CI, if the hypothesis of Normality of errors is not verified.
         \item For asymmetric distributions, the authors should be careful not to show in tables or figures symmetric error bars that would yield results that are out of range (e.g. negative error rates).
         \item If error bars are reported in tables or plots, The authors should explain in the text how they were calculated and reference the corresponding figures or tables in the text.
     \end{itemize}

 \item {\bf Experiments compute resources}
     \item[] Question: For each experiment, does the paper provide sufficient information on the computer resources (type of compute workers, memory, time of execution) needed to reproduce the experiments?
     \item[] Answer: \answerYes{}
     \item[] Justification: We will provide this information in the appendix.
    
 \item {\bf Code of ethics}
     \item[] Question: Does the research conducted in the paper conform, in every respect, with the NeurIPS Code of Ethics \url{https://neurips.cc/public/EthicsGuidelines}?
     \item[] Answer: \answerYes{}
     \item[] Justification: Our research conforms with the Code of Ethics.

 \item {\bf Broader impacts}
     \item[] Question: Does the paper discuss both potential positive societal impacts and negative societal impacts of the work performed?
     \item[] Answer: \answerNA{}
     \item[] Justification: The work has no immediate societal impact.
    
 \item {\bf Safeguards}
     \item[] Question: Does the paper describe safeguards that have been put in place for responsible release of data or models that have a high risk for misuse (e.g., pretrained language models, image generators, or scraped datasets)?
     \item[] Answer: \answerNA{}
     \item[] Justification: No data from the paper has high risk of misuse.

 \item {\bf Licenses for existing assets}
     \item[] Question: Are the creators or original owners of assets (e.g., code, data, models), used in the paper, properly credited and are the license and terms of use explicitly mentioned and properly respected?
     \item[] Answer: \answerYes{}
     \item[] Justification: We cite the creators and comply with the license of all assets used in the paper.

 \item {\bf New assets}
     \item[] Question: Are new assets introduced in the paper well documented and is the documentation provided alongside the assets?
     \item[] Answer: \answerYes{}
     \item[] Justification: We will release the datasets we created.

 \item {\bf Crowdsourcing and research with human subjects}
     \item[] Question: For crowdsourcing experiments and research with human subjects, does the paper include the full text of instructions given to participants and screenshots, if applicable, as well as details about compensation (if any)? 
     \item[] Answer: \answerNA{}
     \item[] Justification: There are no experiments with human subjects.

 \item {\bf Institutional review board (IRB) approvals or equivalent for research with human subjects}
     \item[] Question: Does the paper describe potential risks incurred by study participants, whether such risks were disclosed to the subjects, and whether Institutional Review Board (IRB) approvals (or an equivalent approval/review based on the requirements of your country or institution) were obtained?
     \item[] Answer: \answerNA{}
     \item[] Justification: There are no experiments with human subjects.

 \item {\bf Declaration of LLM usage}
     \item[] Question: Does the paper describe the usage of LLMs if it is an important, original, or non-standard component of the core methods in this research? Note that if the LLM is used only for writing, editing, or formatting purposes and does not impact the core methodology, scientific rigorousness, or originality of the research, declaration is not required.
     this research? 
     \item[] Answer: \answerNA{}
     \item[] Justification: We did not use LLMs to conduct this research.

 \end{enumerate}

\newpage
\appendix

\section{Random Non-Canonical Tokenization}\label{sec:appendix_random_segmentation}

In this section, we provide our algorithm for producing a random non-canonical tokenization, and a proof that each non-canonical tokenization that is \textit{more fine-grained} than the canonical one has equal probability of being output.

\subsection{Algorithm}

We will split each token in a canonical tokenization into smaller tokens (that each exists in the tokenizer's vocabulary). 
We formulate our problem as: Given a valid token \(t\), and a set of vocabulary \(\mathcal V\), construct a sequence of tokens \(seq\) using tokens that exist in \(\mathcal V\) and form \(t\) when concatenated together. We produce \(seq\) using a recursive algorithm. Since there can be many possible \(seq\) for each \(t\), we need to randomly choose one and guarantee that each possible \(seq\) is chosen with equal probability. We achieve this by considering recursion as producing a tree. Each path down the tree corresponds to one possible way to segment \(t\). Each node of the tree represents a segmentation state where we have chosen some number of sub-tokens. At each node, we weigh the choice of which child node to visit by the number of leaves in the sub-tree that is rooted at each child node. This guarantees that each path down the tree is chosen with equal probability since the number of paths down a tree is equal to the number of leaf nodes in that tree. The pseudocode for the algorithm is in \autoref{alg:rts}. 

\subsection{Proof}
\textbf{Goal: } To prove that the random segmentation algorithm chooses one valid segmentation from all possible valid segmentations with uniform probability.

\textbf{Notation: } Let \(W(i)\) denote the number of valid segmentation completions (i.e., the number of leaves in the recursive tree) for the substring starting at index \(i\). In particular, \(W(|token|) = 1\). Note that \(W(i)\) is calculated by the memoized recursive function \(countSegments(i)\), which calculates the number of leaves of the subtree rooted at \(i\).

\textbf{Base Case: } Consider the node corresponding to \(i = |token|\) (the end of the token). Here, there is exactly one valid segmentation (the empty segmentation), so the algorithm returns it with probability 1. That is, every segmentation (in this case, the only one) is chosen with probability
\[
\frac{1}{W(|token|)} = \frac{1}{1} = 1.
\]
Thus, the base case holds.

\textbf{Inductive Hypothesis: } Assume that for any node corresponding to an index \(j\) with \(j > i\) (i.e., deeper in the recursion tree), every complete segmentation (leaf) in the subtree rooted at \(j\) is chosen with probability
\[
\frac{1}{W(j)}.
\]

\textbf{Inductive Step: } Now consider a node corresponding to index \(i\) (with \(i < |token|\)). Suppose that from \(i\) there are \(k\) valid branches corresponding to choosing substrings that end at indices \(j_1, j_2, \dots, j_k\), where for each \(j\) we have \(i < j \leq |token|\) and the substring \(token[i:j]\) is in the vocabulary. By definition,
\[
W(i) = \sum_{r=1}^{k} W(j_r).
\]
The algorithm selects the branch from \(i\) to a specific child \(j\) with probability
\[
P(i \rightarrow j) = \frac{W(j)}{W(i)}.
\]
Once branch \(i \rightarrow j\) is chosen, by the inductive hypothesis every complete segmentation (leaf) in the subtree rooted at \(j\) is chosen with probability
\[
\frac{1}{W(j)}.
\]
Thus, the probability \(P(S)\) of obtaining a particular complete segmentation \(\mathcal V\) that starts at \(i\) by first taking the branch \(i \rightarrow j\) and then following a specific path in the subtree rooted at \(j\) is
\[
P(S) = \frac{W(j)}{W(i)} \cdot \frac{1}{W(j)} = \frac{1}{W(i)}.
\]
Since the factor \(W(j)\) cancels, the probability \(P(S)\) is independent of the particular child \(j\) chosen.

\textbf{Conclusion: } By the inductive step, every complete segmentation (leaf) in the subtree rooted at any index \(i\) is chosen with probability \(\frac{1}{W(i)}\). In particular, when \(i = 0\) (the start of the token), every valid segmentation of the entire token is selected with uniform probability \(\frac{1}{W(0)}\). This completes the proof.


\begin{algorithm}[H]
\caption{Random Token Segmentation \label{alg:rts}}
\begin{algorithmic}[1]
\Function{countSegments}{$start$} \Comment{Cached (using memoization)}
    \If{$start = |token|$}
        \State \Return $1$ \Comment{Reached end; valid segmentation}
    \EndIf
    \State $total \gets 0$
    \For{$end \gets start+1$ \textbf{to} $|token|$}
        \State $substring \gets token[start:end]$
        \If{$substring \in vocabulary$}
            \State $total \gets total + \Call{countSegments}{end}$
        \EndIf
    \EndFor
    \State \Return $total$
\EndFunction

\Function{buildSegments}{$start$}
    \If{$start = |token|$}
        \State \Return $\varnothing$ \Comment{Empty segmentation}
    \EndIf
    \State $validSegments \gets []$
    \State $weights \gets []$
    \For{$end \gets start+1$ \textbf{to} $|token|$}
        \State $substring \gets token[start:end]$
        \If{$substring \in vocabulary$}
            \State $segCount \gets \Call{countSegments}{end}$
            \If{$segCount > 0$}
                \State Append $substring$ to $validSegments$
                \State Append $segCount$ to $weights$
            \EndIf
        \EndIf
    \EndFor
    \If{$validSegments$ is empty}
        \State \Return $\varnothing$
    \EndIf
    \State $chosenSegment \gets$ weightedRandomChoice$(validSegments, weights)$
    \State \Return $[chosenSegment] \; \Vert \; \Call{buildSegments}{start + |chosenSegment|}$
    \Comment{Concatenate chosen segment with segmentation of remaining token}
\EndFunction

\Procedure{SegmentToken}{$token, vocabulary$}
    \If{$\Call{countSegments}{0} = 0$}
        \State \Return $\varnothing$ \Comment{No valid segmentation exists}
    \Else
        \State \Return \Call{buildSegments}{0}
    \EndIf
\EndProcedure
\end{algorithmic}
\end{algorithm}

\section{Evaluation Details}

\subsection{General benchmarks}
For short-answer benchmarks, the system prompt is:
\begin{verbatim}
You are a helpful assistant. 
\end{verbatim}
For multiple-choice benchmarks, the system prompt is:
\begin{verbatim}
You are a helpful assistant. For the following multiple choice questions, 
return the answer only, without any additional reasoning or explanation.
\end{verbatim}
\label{gen_bchmk_eval_dtls}


\paragraph{MATH}
MATH is a dataset composed of fairly difficult, competition level math problems \cite{hendrycks-etal-2021-measuring}. The test set is composed of short answer problem that describe some scenario and asks the model to output a mathematically correct answer. 

\paragraph{GSM8K}
GSM8K is a dataset consisting of relatively simple math questions that would appear in grade school math exams \cite{cobbe2021trainingverifierssolvemath}. For GSM8K, the evaluations were done in the same manner as MATH. 

\paragraph{MMLU}
MMLU is a benchmarks comprising of multiple choice questions from a wide variety of subjects. \cite{hendrycks2021measuring} We sampled 500 questions from MMLU for our evaluation. We instructed the model to only output one answer to each question without any explanation.  

\paragraph{Alpaca Eval}
Alpaca Eval is an evaluation benchmark where generations from language models against given prompts are compared and judged by an annotator model. \cite{dubois2023alpacafarm} The metric used was raw winrate of the perturbed model as judged by a language model. The annotator we used was \textit{alpaca\_eval\_gpt4}, which has been shown to have the highest Spearman and Pearson correlation coefficient with human annotators.

\paragraph{ARC Challenge and ARC Easy}
Contains multiple choice questions with four options each, taken from grade school science exams \cite{clark2018thinksolvedquestionanswering}. ARC Easy is tests basic science knowledge while ARC Challenge requires some procedural reasoning. 

\paragraph{BoolQ} Contains true or false questions along with a context passage that provides the answer to the question. \cite{clark-etal-2019-boolq}

\paragraph{CommonsenseQA} Contains multiple choice questions with five options each that requires common sense knowledge to answer. \cite{talmor-etal-2019-commonsenseqa}

\paragraph{COPA} Contains multiple choice questions with two options each that tests knowledge of cause and effect. \cite{roemmele_choice_2011}

\paragraph{CUTE} Contains questions that require the model to manipulate sentence-level, word-level, and character-level structure for strings. \citep{edman-etal-2024-cute}

\paragraph{DROP} contains questions that potentially require reasoning multiple pieces of information present in a given passage. \cite{dua-etal-2019-drop}

\paragraph{HellaSwag} contains multiples choice questions with four options each that asks for the most natural continuation to some given context. \cite{zellers-etal-2019-hellaswag}

\paragraph{JeopardyQA} contains short answer questions from the ``Jeopardy!" game show. \cite{jeopardy2020}

\paragraph{OpenbookQA} contains multiple choice questions with four options each that require some multi-step and common sense reasoning. \cite{mihaylov2018suitarmorconductelectricity}

\paragraph{PIQA} contains multiple choice questions that require reasoning about the physical world. \cite{bisk-etal-2020-piqa}

\paragraph{TriviaQA} contains short answer questions that requires knowledge of the world. \cite{joshi2017triviaqalargescaledistantly}

\paragraph{Winograd} contains multiple choice questions with two options that asks to determine what a pronoun might refer to. Answering these questions require knowledge of commen sense and surrounding context. \cite{levesque-etal-2012-winograd}

\paragraph{Winogrande} contains questions in the same format of Winograd but there are more questions and the questions are harder. \cite{sakaguchi-etal-2021-winogrande}

\paragraph{TOFU} contains general short answer questions that tests the model's ability to process world knowledge. This is the retain set of the task of fictitious unlearning dataset. 
\cite{maini2024tofutaskfictitiousunlearning}

\paragraph{WikidataQA} require models to complete factual statements. \citep{srivastava2023beyond}


\subsection{Constructed Benchmarks}

\begin{table}[t!]
    \centering\small
    \caption{System prompt for tasks in \autoref{sec:better_tokenizations}. See \autoref{tab:example_prompts} for example instructions.}
    \vspace{1em}
    \begin{tabular}{p{380pt}}
    \toprule
        \textbf{Counting characters:} \texttt{You are a helpful assistant. The following prompt will ask you to return a 
        sequence of words. Only return the sequence, separated by spaces. Do not 
        provide any additional text or explanation.} \\\midrule
        \textbf{Code Description:} 
        \texttt{You are a programming assistant trained to analyze and interpret code snippets. When provided with a code snippet and a set of answer choices (A, B, C, or D), your task is to evaluate the code, determine its behavior, and select the answer that best describes this behavior. Your response must be a single letter: A, B, C, or D. Do not provide explanations or additional text unless explicitly requested.}\\\midrule
        \textbf{Arithmetic:} \texttt{You are a computational assistant trained to evaluate arithmetic operations. When provided with an arithmetic expression, calculate the result and round it to the nearest integer. Respond only with the rounded result, without any additional text or explanation.} \\
        \bottomrule
    \end{tabular}
    \label{tab:system_example_prompts}
    \vspace{-1em}
\end{table}

In this section, we provide more detail on how datasets we use in \autoref{sec:better_tokenizations} are constructed.

\label{subsec:perf_improve}
\paragraph{Count Characters Task}
The prompt asks the model to count the number of occurrences of a given character in a 10-character word; we always use the most frequently occurring character.
Evaluation was done, similar to GSM and MATH, by finding the last number in the generated response. 
Generations without any numbers are considered incorrect. 
\paragraph{Generate Acronym Task}
The model is asked to generate a sequence of words whose first letters form a randomly sampled five character string. 
For evaluation, we take the first character of each whitespace-delimited word and check if it matches the desired acronym. 
\paragraph{Codeline Description Task}
The model is asked to comprehend a piece of code and choose the best description from four options. 
\paragraph{Arithmetic Task} 
The model is asked to perform addition or subtraction with 10 digit numbers. We use regex to extract numbers from the generation, which are then compared to the ground truth answer.

\subsection{Metrics of generation quality}
Here we provide additional details on the metrics defined in \autoref{subsec:when}.

\paragraph{Spelling}
We use the top 10000 most frequently appearing English words in Google's trillion word corpus. 
We only consider words with more than one character.
This is because sometimes base models will repeatedly generate the same letter, and since all English letters are in the word list, the generation would receive a high score.

\paragraph{Grammaticality}
One drawback with this evaluation method is that oftentimes the model would repeat the same letter over and over again, or start counting numbers. 
In both of these cases, there are no detected grammar mistakes, however they are still obviously gibberish. Therefore, we only calculate grammaticality scores for generations that receive a score $\geq 0.5$ on spelling; otherwise, we give it a grammaticality score of 0. 

\paragraph{Win rate} 
Similar to evaluation in \autoref{sec:experiments}, we also used \texttt{alpaca\_eval\_gpt4} as the evaluator and report raw win rate. 
In \ref{subsec:when}, the win rate is calculated against generations conditioned on input with canonical tokenization. In \ref{subsec:ablations}, the win rate is against generations from the \textbf{No Ablation} setting when also given character-level tokenization.
By construction, the win rate of the \textbf{No Ablation} setting itself is 50\%.


\begin{table}[t!]
    \centering\small
    \caption{Data format of ablations in \autoref{subsec:ablations}.}
    \vspace{1em}
    \begin{tabular}{p{380pt}}
    \toprule
        \textbf{No ablation:}
        \texttt{<|user|>Provide a detailed analysis of Candace Parker's defensive techniques in her recent games, excluding the words "aggressive" and "blocking", in the format of a sports commentary script. <|assistant|>[Sports Commentary Script]}\\
        \texttt{[Opening Scene...}
        \\\midrule
        \textbf{QA Template:} \texttt{Question: Provide a detailed analysis of Candace Parker's defensive techniques in her recent games, excluding the words "aggressive" and "blocking", in the format of a sports commentary script. Answer: [Sports Commentary Script]}\\
        \texttt{[Opening Scene...} \\\midrule
        \textbf{Removing the chat template:} \texttt{Provide a detailed analysis of Candace Parker's defensive techniques in her recent games, excluding the words "aggressive" and "blocking", in the format of a sports commentary script. [Sports Commentary Script]}\\
        \texttt{[Opening Scene...}\\\midrule
        \textbf{Removing the instruction:} 
        \texttt{<|user|>[Sports Commentary Script]}\\
        \texttt{[Opening Scene: A packed basketball arena, with fans eagerly awaiting the analysis of Candace Parker’s recent performances on the court.]}\\
        \texttt{Commentator 1: Welcome back, basketball fans! <|assistant|>Tonight, we're diving into the defensive prowess of Candace Parker...} \\
        \bottomrule
    \end{tabular}
    \label{tab:ablation_data_formats}
    \vspace{-1em}
\end{table}

\subsection{Ablation Settings}

For ablations on the data format, see examples of formatted data in \autoref{tab:ablation_data_formats}.
Our finetuning code was forked from \href{https://github.com/allenai/open-instruct}{\texttt{allenai/open-instruct}}. The exact finetune recipe is given below:
\begin{itemize}
    \item Setup: 8 L40S GPUs
    \item Gradient Accumulation Steps: 20
    \item Per Device Train Batch Size: 2
    \item Mixed Predision: bf16
    \item Max Seq Length: 4096
    \item Learning Rate: 5e-06
    \item LR Scheduler Type: Linear
    \item Warmup Ratio: 0.03
    \item Weight Decay: 0
    \item Epochs: 1
    \item Seed: 123
\end{itemize}






\subsection{Disentangling understanding from generation}

For these tasks, we use 500 words randomly sampled from Google's 10000 English word list\footnote{\url{https://github.com/first20hours/google-10000-english/blob/master/google-10000-english.txt}}.

\paragraph{Word Repeat} 
An example prompt is shown below.

\begin{verbatim}
Repeat each word directly, while correcting any typos.

Question: guarantees
Answer: guarantees

Question: revelation (character-level tokenization)
Answer: 
\end{verbatim}

\paragraph{Identifying Misspellings}
We obtain the misspelled word by randomly adding, removing, or substituting a single character from the word.
An example prompt is shown below.

\begin{verbatim}
Question: Which of the two words contains a misspelling? Respond directly 
with the answer option.

Question: 

A. guarantees 
B. garantees

Answer: B

{9 more in context examples}

Question:

A. farmer (character-level tokenization)
B. farme (canonical tokenization)
\end{verbatim}

\section{Additional Results}
\subsection{Evaluation on Chinese Benchmarks}
We also investigate how robust language models are given character-level tokenization of text in Chinese as evaluated on two tasks, Chinese GSM \cite{shi2022languagemodelsmultilingualchainofthought} (part of multilingual GSM benchmarks) and Chinese MMLU \cite{li-etal-2024-cmmlu}.
Note that since each Chinese character is usually represented with three bytes under UTF-8 encoding, this is not equivalent to byte-level tokenization. 
We focus on \lm{Qwen-2.5-7B-Instruct} as \lm{Llama-3.1-8B-Instruct} and \lm{OLMo-2-7B-Instruct} do not officially support Chinese.
As shown in \autoref{fig:chinese-results}, we observe a similar robustness on Chinese, with performance dropping by only $\sim3\%$ on each task.

\begin{table}[t]
    \centering\small
    \caption{\lm{Qwen-2.5-7B-Instruct} is robust to character-level tokenization of Chinese text.}
    \vspace{1em}
    \begin{tabular}{lcc}
        \toprule
        Benchmark & Canon & Char \\\midrule
        Chinese MMLU & 77.8 & 74.2 \\
        Chinese GSM  & 78.8 & 76.8 \\
        \bottomrule
    \end{tabular}
    \label{fig:chinese-results}
\end{table}

\end{document}

%% file: alisa_macros.tex
\usepackage{amsmath}
\usepackage{wrapfig}
\usepackage{makecell}
\usepackage{multicol}
\usepackage{multirow}
\usepackage{caption}
\usepackage{xcolor}
\usepackage{xspace}
\usepackage{graphicx}
\usepackage{listings}
\usepackage{longtable}
\usepackage{afterpage}
\usepackage{tabularray}
\usepackage{booktabs,arydshln}
\UseTblrLibrary{booktabs}
\usepackage{graphicx}
\usepackage{amsmath}
\usepackage{caption}
\usepackage{subcaption}
\usepackage{xspace}
\usepackage{array}
\usepackage{comment}

\definecolor{gray}{HTML}{F0F0F0}
\makeatletter
\renewcommand{\sectionautorefname}{\S\@gobble}
\renewcommand{\subsectionautorefname}{\S\@gobble}
\renewcommand{\subsubsectionautorefname}{\S\@gobble}
\renewcommand{\appendixautorefname}{Appendix \@gobble}
\makeatother

\newcommand{\lm}[1]{\textsc{#1}}

\definecolor{purp}{HTML}{791f87}

\newcommand{\aspace}{\hspace{1em}}
\newcommand{\uw}{$^{\heartsuit}$}
\newcommand{\aiTwo}{$^{\clubsuit}$}
\newcommand{\stanford}{$^{\spadesuit}$}

\makeatletter
\def\adl@drawiv#1#2#3{%
    \hskip.5\tabcolsep
    \xleaders#3{#2.5\@tempdimb #1{1}#2.5\@tempdimb}%
            #2\z@ plus1fil minus1fil\relax
    \hskip.5\tabcolsep}
\newcommand{\cdashlinelr}[1]{%
\noalign{\vskip\aboverulesep
       \global\let\@dashdrawstore\adl@draw
       \global\let\adl@draw\adl@drawiv}
\cdashline{#1}
\noalign{\global\let\adl@draw\@dashdrawstore
       \vskip\belowrulesep}}
\makeatother

\newcommand{\whitespace}{\,\textvisiblespace\,}